\documentclass{llncs}

\usepackage{times}

\usepackage{latexsym}
 
\usepackage{graphicx}
\usepackage{wrapfig}

\usepackage{xspace}

\begin{document}

\title{SAT vs CSP: a commentary\thanks{The original CP 2000 paper is 
available without pay-wall at https://tinyurl.com/SATvCSP}}

\author{Toby Walsh}
\institute{TU Berlin and UNSW Sydney\\
Germany and Australia\\
tw@cse.unsw.edu.au}

\maketitle

\begin{abstract}
In 2000, I published a relatively comprehensive study of mappings between
propositional satisfiability (SAT) and 
constraint satisfaction problems (CSPs)
\cite{wcp2000}. I analysed four different mappings of SAT problems into CSPs,
and two of CSPs into SAT problems. For each mapping, I compared the
impact of achieving arc-consistency on the CSP with unit propagation
on the corresponding SAT problems, and lifted these results to CSP algorithms
that maintain (some level of ) arc-consistency during search like FC and
MAC, and to the Davis-Putnam procedure (which performs unit
propagation at each search node). These results helped provide some insight into the relationship
between propositional satisfiability and constraint satisfaction that
set the scene for an important and valuable body of work that
followed. I discuss here what prompted the paper, and what followed. 
\end{abstract}

\newcommand{\DLex}{\mbox{\sc DoubleLex}}
\newcommand{\snakelex}{\mbox{\sc SnakeLex}}

\newcommand{\set}{\mathcal}
\newcommand{\myset}[1]{\ensuremath{\mathcal #1}}

\renewcommand{\theenumii}{\alph{enumii}}
\renewcommand{\theenumiii}{\roman{enumiii}}
\newcommand{\figref}[1]{Figure \ref{#1}}
\newcommand{\tref}[1]{Table \ref{#1}}
\newcommand{\myldots}{\ldots}

\newtheorem{myproblem}{Problem}
\newtheorem{mydefinition}{Definition}
\newtheorem{mytheorem}{Proposition}
\newtheorem{mylemma}{Lemma}
\newtheorem{myexample}{Running Example}{\bf}{\it}
\newtheorem{mytheorem1}{Theorem}
\newcommand{\myproof}{\noindent {\bf Proof:\ \ }}
\newcommand{\myqed}{\mbox{$\Box$}}
\newcommand{\myend}{\mbox{$\clubsuit$}}

\newcommand{\mymod}{\mbox{\rm mod}}
\newcommand{\mymin}{\mbox{\rm min}}
\newcommand{\mymax}{\mbox{\rm max}}
\newcommand{\range}{\mbox{\sc Range}}
\newcommand{\roots}{\mbox{\sc Roots}}
\newcommand{\myiff}{\mbox{\rm iff}}
\newcommand{\alldifferent}{\mbox{\sc AllDifferent}}
\newcommand{\permutation}{\mbox{\sc Permutation}}
\newcommand{\disjoint}{\mbox{\sc Disjoint}}
\newcommand{\cardpath}{\mbox{\sc CardPath}}
\newcommand{\CARDPATH}{\mbox{\sc CardPath}}
\newcommand{\common}{\mbox{\sc Common}}
\newcommand{\uses}{\mbox{\sc Uses}}
\newcommand{\lex}{\mbox{\sc Lex}}
\newcommand{\usedby}{\mbox{\sc UsedBy}}
\newcommand{\nvalue}{\mbox{\sc NValue}}
\newcommand{\slide}{\mbox{\sc CardPath}}
\newcommand{\sliden}{\mbox{\sc AllPath}}
\newcommand{\SLIDE}{\mbox{\sc CardPath}}
\newcommand{\circularslide}{\mbox{\sc CardPath}_{\rm O}}
\newcommand{\among}{\mbox{\sc Among}}
\newcommand{\mysum}{\mbox{\sc MySum}}
\newcommand{\amongseq}{\mbox{\sc AmongSeq}}
\newcommand{\atmost}{\mbox{\sc AtMost}}
\newcommand{\atleast}{\mbox{\sc AtLeast}}
\newcommand{\element}{\mbox{\sc Element}}
\newcommand{\gcc}{\mbox{\sc Gcc}}
\newcommand{\gsc}{\mbox{\sc Gsc}}
\newcommand{\contiguity}{\mbox{\sc Contiguity}}
\newcommand{\PRECEDENCE}{\mbox{\sc Precedence}}
\newcommand{\assignnvalues}{\mbox{\sc Assign\&NValues}}
\newcommand{\linksettobooleans}{\mbox{\sc LinkSet2Booleans}}
\newcommand{\domain}{\mbox{\sc Domain}}
\newcommand{\symalldiff}{\mbox{\sc SymAllDiff}}
\newcommand{\alldiff}{\mbox{\sc AllDiff}}

\newcommand{\slidingsum}{\mbox{\sc SlidingSum}}
\newcommand{\MaxIndex}{\mbox{\sc MaxIndex}}
\newcommand{\REGULAR}{\mbox{\sc Regular}}
\newcommand{\regular}{\mbox{\sc Regular}}
\newcommand{\precedence}{\mbox{\sc Precedence}}
\newcommand{\STRETCH}{\mbox{\sc Stretch}}
\newcommand{\SLIDEOR}{\mbox{\sc SlideOr}}
\newcommand{\NAE}{\mbox{\sc NotAllEqual}}
\newcommand{\mytheta}{\mbox{$\theta_1$}}
\newcommand{\mysigma}{\mbox{$\sigma_2$}}
\newcommand{\mysigmatwo}{\mbox{$\sigma_1$}}

\newcommand{\todo}[1]{{\tt (... #1 ...)}}
\newcommand{\myOmit}[1]{}
\newcommand{\nina}[1]{#1}
\newcommand{\ninacp}[1]{#1}

\newcommand{\dpsb}{DPSB}

\section{Introduction}

Two decades after it was written, a paper I wrote comparing
propositional satisfiability and constraint satisfaction
\cite{wcp2000} has,
to my surprise, become one of the most cited works from the CP 2000 conference
(with 353 cites currently on Google Scholar). Indeed, to my surprise, 
it currently ranks second in citations of {\em all} of  my scientific papers. 
In part, this is luck. There were a number of encodings
of CSPs into SAT that were well known but hadn't been
documented extensively in prior work. 
It was also luck that had me write these down as I nearly
didn't submit the paper since I remember thinking that many of its results
were not very deep. And it was luck that I had started
working in the constraint programming community
as a result of having come from the area of propositional satisfiability. 
This put me in the natural place to draw the connections
between the two areas. It was obvious to me that 
arc consistency and unit propagation were making 
very similar inferences. But perhaps this seemed less immediate
at that time to people without feet in both areas. This paper
attempted to correct this. 

\section{Contributions}

This wasn't the first paper to look at mappings between SAT and 
CSPs. For instance, Bennaceur had previously looked at encoding SAT problems as CSPs
\cite{bennaceur1}, whilst 
 G\'{e}nisson and J\'{e}gou had looked at encoding CSPs as SAT problems 
\cite{genisson1}.
However,  earlier studies like these has only considered a limited
number of mappings.
This was perhaps the first study to attempt to look more comprehensively
about mappings between the two domains, and certainly to do so
in both directions at once. 
The paper considered 4 encodings of SAT into CSPs
(dual, hidden variable, literal and the non-binary encoding),
2 encodings of CSPs into SAT (direct and log encoding),
and studied some of the theoretical properties of these 
mappings.

To be more specific, I proved that achieving arc-consistency on the dual
encoding does more work than unit propagation on the original SAT problem,
whilst achieving arc-consistency on the hidden variable and literal encodings
does essentially the same work. I then extended these results to algorithms
that maintain some level of arc-consistency during search like
forward checking (FC) and maintaining arc-consistency (MAC), and
DP which performs unit propagation at each search node supposing
equivalent
branching heuristics. I proved that the Davis Putnam (DP) algorithm strictly dominates
FC applied to the dual encoding, is incomparable to MAC applied to the dual
encoding, explores the same number of branches as MAC applied to the hidden
variable encoding, and strictly dominates MAC applied to the literal encoding.
Finally I proved that unit propagation on the direct encoding
does less work than achieving arc-consistency on the original problem, but more
work than unit propagation on the log encoding. DP on the direct encoding
explores the same size search tree as FC applied to the original problem, but is
strictly dominated by MAC. By comparison, DP on the log encoding is strictly
dominated by both FC and MAC applied to the original problem.
A follow up 
paper by my colleague, Ian Gent looked
at the closely related support encoding of CSPs
into SAT \cite{gecai2002}. As the name suggest, this encodes not 
conflicts (as the dual and hidden-variable encodings do), but
supports. 

\section{What happened next?}

Though these results were not deep, they did set the 
scene for a range of future research that brought
ideas and techniques from propositional 
satisfiability into constraint satisfaction. Interestingly,
the direction has been almost all one way. I am not
aware of much work exporting ideas and techniques
from constraint satisfaction into propositional
satisfiability. I leave it as an interesting open
question for the reader as to why this might be
the case. 

\subsection{Solving CSPs by encoding into SAT}

The most immediate and direct application of the 
encodings discussed in \cite{wcp2000} and \cite{gecai2002} is to 
encode CSPs into SAT so we can bring to 
bear the advanced and powerful SAT solvers based on
complete and local search methods that are 
now available to download. 
For instance, we showed that it was effective to 
encode global {\sc Grammar} and {\sc Regular}
constraints into SAT \cite{qwcp07}, and gave state of the art 
results on some shift rostering problems. 
Similar encodings of CSPs into answer set programs (ASPs)
\cite{dwiclp10,dwijcai11,dwiclp11} and into 
integer linear programs (ILPs) \cite{hkwaor2003}
have also been shown to be highly effective. 
Many medal winners in the MiniZinc constraint solving
competition have used SAT encodings (e.g. PicatSAT)
or SAT inspired algorithms (e.g. HaifaCSP and MinisatID).

\subsection{Lazy clause generation}

Perhaps the most important work that followed is
the research of
Ohrimenko, Stuckey, Codish and colleagues on
lazy clause generation \cite{lazyclause}. 
This shows that domain propagation on a CSP can be achieved by 
generating clauses for a SAT solver. 
Since a naive static translation is impractical except
in limited cases, the clauses are generated lazily
as needed. Such solving techniques open up many
new possibilities like nogood learning and the use of SAT
based branching heuristics. Lazy clause solvers remain
today some of the fastest on many classes of challenging problems. 

\subsection{Analysing global constraints}

The final area I want to mention is one that hasn't received
the attention I believe it deserves since it touches on one of the
aspects of constraint solving that makes it special: namely, global
constraints. Encodings of constraints into SAT have played a central role
in understanding how far we can effectively decompose global
constraints. There are a number of promising 
results decomposing global constraints
into smaller, and easier to propagate
constraints (e.g. 
\cite{gswaij2000,gswercrim2000,qwcp07,bnqswcp07,bknqwijcai09,bknqwcp10}). 
However, by drawing on lower bound results from circuit complexity, 
we proved some fundamental limits to
such decompositions \cite{bknwijcai09}. 

Consider encodings of global
constraints into propositional formulae in 
conjunctive normal form that have  the fundamental 
property that unit propagation on the decomposition enforces the same level of consistency as a
specialized propagation algorithm like that which maintains 
arc consistency. We proved that
a constraint propagator has a a polynomial size decomposition if and only if it can be computed by a
polynomial size monotone Boolean circuit. Lower
bounds on the size of monotone Boolean circuits
thus translate to lower bounds on the size of decompositions of global constraints. For instance,
we proved that there is no polynomial sized decomposition of the domain consistency propagator for
the {\sc AllDifferent} constraint.
On the other hand, bound and range consistency propagators of 
{\sc AllDifferent} can be decomposed effectively into SAT, as we demonstrate in
\cite{bknqwijcai09}.

\section{Conclusions}

Constraint satisfaction and propositional satisfiability
are closely related areas. Mappings between the two
have proved a fruitful area for research in the last two
decades. Somewhat surprisingly, it has mostly been 
an one-way flow, with tools and techniques from
propositional satisfiability being used to 
help solve constraint satisfaction problems. It's interesting
to consider if we could not also look at the reverse
direction?

\section*{Acknowledgments}

This research was 
funded by the European Research Council under the Horizon 2020 Programme via AMPLify
670077.

\bibliographystyle{alpha}

\begin{thebibliography}{BKNW10}

\bibitem[Ben96]{bennaceur1}
H.~Bennaceur.
\newblock The satisfiability problem regarded as a constraint satisfaction
  problem.
\newblock In W.~Wahlster, editor, {\em Proceedings of the 12th ECAI}, pages
  155--159. European Conference on Artificial Intelligence, Wiley, 1996.

\bibitem[BKN{\etalchar{+}}09]{bknqwijcai09}
C.~Bessiere, G.~Katsirelos, N.~Narodytska, C-G. Quimper, and T.~Walsh.
\newblock Decompositions of all different, global cardinality and related
  constraints.
\newblock In {\em Proceedings of 21st IJCAI}, pages 419--424. International
  Joint Conference on Artificial Intelligence, 2009.

\bibitem[BKNW09]{bknwijcai09}
C.~Bessiere, G.~Katsirelos, N.~Narodytska, and T.~Walsh.
\newblock Circuit complexity and decompositions of global constraints.
\newblock In {\em Proceedings of 21st IJCAI}, pages 412--418. International
  Joint Conference on Artificial Intelligence, 2009.

\bibitem[BKNW10]{bknqwcp10}
C.~Bessiere, G.~Katsirelos, N.~Narodytska, and C-G. Quimper~T. Walsh.
\newblock Decomposition of the {NValue} constraint.
\newblock In D.~Cohen, editor, {\em Proceedings of the 16th International
  Conference on the Principles and Practice of Constraint Programming (CP
  2010)}, volume 6308 of {\em Lecture Notes in Computer Science}. Springer,
  2010.

\bibitem[BNQ{\etalchar{+}}07]{bnqswcp07}
S.~Brand, N.~Narodytska, C.-G. Quimper, P.~Stuckey, and T.~Walsh.
\newblock Encodings of the sequence constraint.
\newblock In {\em 13th International Conference on Principles and Practices of
  Constraint Programming (CP-2007)}. Springer-Verlag, 2007.

\bibitem[DW10]{dwiclp10}
C.~Drescher and T.~Walsh.
\newblock A translational approach to constraint answer set solving.
\newblock {\em Theory and Practice of Logic Programming}, 10(4-6):465--480,
  2010.

\bibitem[DW11a]{dwiclp11}
C.~Drescher and T.~Walsh.
\newblock Modelling {GRAMMAR} constraints with answer set programming.
\newblock In {\em Proceedings of the 27th International Conference on Logic
  Programming (ICLP 2011)}, 2011.

\bibitem[DW11b]{dwijcai11}
C.~Drescher and T.~Walsh.
\newblock Translation-based constraint answer set solving.
\newblock In {\em Proceedings of the 22nd International Joint Conference on
  Artificial Intelligence (IJCAI-2011)}. International Joint Conference on
  Artificial Intelligence, 2011.

\bibitem[Gen02]{gecai2002}
I.P. Gent.
\newblock Arc consistency in {SAT}.
\newblock In F.~van Harmelen, editor, {\em Proceedings of ECAI-2002}. IOS
  Press, 2002.

\bibitem[GJ96]{genisson1}
R.~Genisson and P.~Jegou.
\newblock {Davis} and {Putnam} were already forward checking.
\newblock In W.~Wahlster, editor, {\em Proceedings of the 12th ECAI}, pages
  180--184. European Conference on Artificial Intelligence, Wiley, 1996.

\bibitem[GSW00a]{gswaij2000}
I.P. Gent, K.~Stergiou, and T.~Walsh.
\newblock Decomposable constraints.
\newblock {\em Artificial Intelligence}, 123(1-2):133--156, 2000.

\bibitem[GSW00b]{gswercrim2000}
I.P. Gent, K.~Stergiou, and T.~Walsh.
\newblock Decomposable constraints.
\newblock In K.~Apt, A.~Kakis, E.~Monfroy, and F.~Rossi, editors, {\em New
  Trends in Constraints}, number 1865 in LNAI, pages 134--149. Springer, 2000.

\bibitem[HKW03]{hkwaor2003}
B.~Hnich, Z.~Kiziltan, and T.~Walsh.
\newblock Hybrid modelling for robust solving.
\newblock {\em Anals of Operations Research}, 130(1-4):19--39, 2003.

\bibitem[OSC09]{lazyclause}
Olga Ohrimenko, Peter~J. Stuckey, and Michael Codish.
\newblock Propagation via lazy clause generation.
\newblock {\em Constraints}, 14(3):357--391, 2009.

\bibitem[QW07]{qwcp07}
C.-G. Quimper and T.~Walsh.
\newblock Decomposing global grammar constraints.
\newblock In {\em 13th International Conference on Principles and Practices of
  Constraint Programming (CP-2007)}. Springer-Verlag, 2007.

\bibitem[Wal00]{wcp2000}
T.~Walsh.
\newblock {SAT} v {CSP}.
\newblock In Rina Dechter, editor, {\em 6th International Conference on
  Principles and Practices of Constraint Programming (CP-2000)}, pages
  441--456. Springer-Verlag, 2000.

\end{thebibliography}

\newcommand{\etalchar}[1]{$^{#1}$}

\end{document}